\documentclass{article}

\usepackage{arxiv}

\usepackage[utf8]{inputenc} 
\usepackage[T1]{fontenc}    
\usepackage{hyperref}       
\usepackage{url}            
\usepackage{booktabs}       
\usepackage{amsfonts}       
\usepackage{nicefrac}       
\usepackage{microtype}      
\usepackage{lipsum}
\usepackage{graphicx}
\usepackage[
backend=biber,
style=numeric,
sorting=ynt
]{biblatex}

\graphicspath{ {./images/} }

\title{Learning, transferring, and recommending performance knowledge with monte carlo tree search and neural networks}

\author{
 Don M. Dini \\
  dmdini.com\\
  Menlo Park, CA\\
  \texttt{don@dmdini.com} \\
}

\begin{document}

\maketitle
\begin{abstract}
Making changes to a program to optimize its performance is an unscalable task that relies entirely upon human intuition and experience. In addition, companies operating at large scale are at a stage where no single individual understands the code controlling its systems, and for this reason, making changes to improve performance can become intractably difficult. In this paper, a learning system is introduced that provides AI assistance for finding recommended changes to a program. Specifically, it is shown how the evaluative feedback, delayed-reward performance programming domain can be effectively formulated via the Monte Carlo tree search (MCTS) framework. It is then shown that established methods from computational games for using learning to expedite tree-search computation can be adapted to speed up computing recommended program alterations. Estimates of expected utility from MCTS trees built for previous problems are used to learn a sampling policy that remains effective across new problems, thus demonstrating transferability of optimization knowledge. This formulation is applied to the Apache Spark distributed computing environment, and a preliminary result is observed that the time required to build a search tree for finding recommendations is reduced by up to a factor of 10x. 
\end{abstract}


\section{Introduction}
For distributed software running in a production environment, it is important to optimize its performance to the degree possible. The task is to make changes to a program, or the environment in which it runs, in order to improve its performance, measured according to metrics such as execution time, memory and disk usage, or a custom defined quantity. Today, this is accomplished via the intuition, knowledge, and experience of the data engineer, and for this reason is completely unscalable; i.e. the only way to solve more problems is to hire more engineers. Additionally, the scale and complexity of solvable problems are limited to those that can fit in a human engineer’s mind. For example, optimizing for multiple metrics (e.g. speed and cost), is usually too complex a task to reason about. Companies operating at large scale are at a stage where no single individual understands the code controlling its systems, and for this reason, making changes to improve performance can become intractably difficult. To move beyond this boundary, we require the assistance of AI.

	In this paper, a learning system is described for performing the following. First, given a distributed software program as input, possible sequences of small changes that affect its performance are explored. The knowledge of which changes were successful, and which were not, is then learned and generalized from in the form of a model. This model then improves the speed with which recommendations are found in future, unseen software programs.
	
	Most related work on optimizing the performance of user submitted programs focuses on interpreting and executing the given instructions in the best possible manner (e.g. optimal query execution). In addition to good parameter settings, a huge bulk of the responsibility for a program’s performance is the direct contents of the program. That is the focus in this work.
	
	In this paper, we argue that since introducing changes to a program is an evaluative feedback task (i.e. one must propose a change to see how it performs), and reward-delayed (several subsequent alterations must often be made, which by themselves may produce little effect), the most natural formulation is a decision theoretic framework. Once cast in the language of decision theory, we show how finding recommended program alterations can be formulated as applying an optimal policy.
	
	In decision theoretic formulations, a domain is resolved into (possibly infinitely many) states, $S = \{ s_1, \cdots \}$, with $s_0$ denoting the initial state.At a given state, the agent can perform an action from the set $ A=\{ a_1, a_2, \cdots\} $. Upon doing so, the agent arrives in a new state in  according to a (known or unknown) probability distribution $p\left( s_j | s_i, a \right)$, and receives a reward according to (known or unknown) distribution $r\left(s_i, a \right)$. A policy is a mapping from states  to actions  (or, more generally, a probably distribution over ). Varying this policy has an impact on how much reward an agent accumulates by acting in the domain over time, and so the task in decision theory is to find an optimal such policy (or at least a good one), in the sense of maximizing long term expected reward. In the performance programming domain, the initial state is the contents of the program, and the action space is comprised of potential valid alterations. A recommendation is obtained by applying a sequence of such alterations, via a policy, to the initial state.
	
	Monte-Carlo Tree Search (MCTS) ~\cite{browne2012survey} is an effective, simple framework for finding such a policy. The MCTS framework works by estimating long-term expected utility via sampling possible rollouts. How the rollouts are sampled has quite a significant impact. For example, in the UCT~\cite{kocsis2006bandit} family of algorithms, The (domain-independent) UCB1 heuristic is applied to focus sampling on the most promising parts of the action space, resulting in great advantage over flat, uniform sampling.
	
	More generally, however, the action space is focused by sampling from a policy distribution, $p\left(a|s \right)$. As in all problems involving reinforcement learning, this policy must be selected in such a way that balances exploration and exploitation. One typically favors exploration early on, and gradually focuses on the actions that have proved fruitful.
	
	This process of learning from the environment does not need to begin from scratch for each problem, however. The successful application of MCTS to computational games has led to established methods for using learned domain information to expedite the search process. Specifically, domain knowledge can be represented in $p\left( a|s \right)$ beforehand via learning. In~\cite{coulom2007computing} for example, sample game records for Go are examined to learn effective policy distributions. It is precisely this same mechanism of speedup via transfer of learned knowledge that we capture and apply to the performance programming domain. Specifically, we use search trees built via UCT algorithm to generate expected-utility data. This data is then used to learn a representation for $p\left(a|s \right)$, which is effective across a range of problems.
	
	We use as a test bed the Apache Spark distributed programming environment~\cite{zaharia2012resilient}. Spark is a popular, large scale distributed computing platform which uses as its core abstraction the resilient distributed dataset (RDD). The action space is comprised primarily of controlling how RDDs are distributed amongst available machines, and how they use memory. In a typical program, the branching factor at any given state is b$\sim$50, and a typical search tree producing a solution has depth d$\sim$5. Although Spark is used as a test bed in this paper, the goal is to help develop a framework that generalizes to AI assisted performance programming more generally.

\section{Related work}
\label{related}
In the problem space of optimizing a user submitted program according to a specified performance metric, there is a large body of work. Once a program has been authored, there is of course much work on optimizing how a system interprets and executes a program. Work along these lines can be classified by the environment in which the program is to be run. For example, there is query execution optimization~\cite{jarke1984query}, as well as automated parallelization in distributed environments~\cite{armbrust2015spark}. From the user’s perspective, the same program simply runs more quickly and more efficiently.

	The behavior of computing frameworks is often governed by tens or hundreds of configurable parameters. In these cases, there is work towards suggesting good values for parameters, such as using the more general framework of Bayesian Optimization~\cite{shahriari2016taking} for model tuning~\cite{clark2015adaptive}. In the case of machine learning models, performance as well as accuracy often depend upon a host of hyper parameters, such as learning rate and architecture. ~\cite{zoph2016neural}, ~\cite{liu2018darts} consider the search for and recommendation of neural architectures. The suggestion of optimal parameters has also been investigated for Apache Spark in particular~\cite{gounaris2018methodology}.
	
	In~\cite{chen2018learning}, the specific problem of optimizing programs consisting of tensor operations is considered. Specifically, given a particular tensor operation, how it is implemented in low level execution has enormous impact in execution performance. The search space of possible programs is large, and transfer learning is employed to gain a reported 2x to 10x speedup.
	
In this paper we consider tunable parameters as fixed, and focus on changing the contents of the program over a well defined action space to improve a user specified, measurable performance metric.

\section{Model description and approach}
\label{model}

\subsection{Performance programming as Monte Carlo Tree Search}
Performance programming, unlike game playing, does not have a long history of MCTS based approaches. Using decision theoretic formulations, and MCTS in particular, is thus not necessarily an immediately obvious choice. Decision theoretic approaches are in fact highly suitable and have much to offer in this problem space. A discussion about why this should be the case is motivated.
	The performance-programming domain is an evaluative feedback domain, in which labeled examples are not available - i.e. one does not typically know beforehand how well a particular alteration will work out. Moreover, once an alteration is selected, one can only evaluate how well this particular choice performed. There is no way of knowing which alteration was in fact the “best”, or even if such an action exists.
	A multi-armed bandit~\cite{sutton2018reinforcement} approach would be appropriate for the above conditions if program alteration could be viewed as a single-stage action. In contrast, performance programming is a delayed-reward environment, making such an approach impossible; a highly impactful alteration must often be supported by other alterations elsewhere in the program, which alone do not clearly contribute an impact.
	States, actions, and rewards can be defined in the performance programming domain in the following manner:

\begin{itemize}
\item $States$. Textual program contents. Extracting features from this content is described in the next section. There are a larger set of features which have an impact on performance. Specifically, there are a large number of run-time parameters~\cite{gounaris2018methodology} which impact performance. In this paper, it is presumed that these parameters are held constant, and program contents alone is considered as a free parameter. In addition to the above, there are factors such as multi-tenancy that make state technically partially-observable.
\item $Actions$. One can produce a program to perform a useful distributed computation in Spark relatively straightforwardly. However, making it perform well in a production environment typically involves painstaking, time-consuming changes to how memory is used, as well as how tasks are distributed among available machines in the cluster. These individual actions are captured here as persisting an individual RDD to memory, adding partitioning to an individual RDD, and lastly, changing joins to broadcast-joins. 
\item $Reward$. In this context reward is considered to be the performance metric as measured in a state. For example, after making an alteration that causes an RDD to be persisted to memory, one can execute the new program, measuring its execution time (or another metric). Technically this can be considered to be a stochastic variable as multi-tenancy as well as other exogenous (and thus difficult to measure) factors can have an impact on a program’s performance.
\end{itemize}

Given this, the problem of finding a recommended alteration to a program to achieve a performance gain, can thus be formulated as finding a policy which optimizes expected reward, where reward is measured as the user’s desired performance metric. 
	Monte-carlo tree search (MCTS) is a highly effective method for doing so. MCTS estimates long term expected utility of starting in a state via sampling rollouts. At each iteration of the UCT family of algorithms, the most fruitful part of the tree to explore is considered, in a manner that balances exploration and exploitation (with provably optimal regret bounds~\cite{auer2002finite}).
	Using the domain-independent UCT algorithm, MCTS is able to reproduce known program-optimization solutions, starting from scratch. For example, given a logically correct, but completely unoptimized logistic regression program, UCT produced, via only the actions in the action space above, an optimized logistic regression with execution time equivalent to the Spark MLLib implementation. The MCTS framework is clearly capable of expressing a solution - the question is the computational effort expended in producing one. 
	In addition, the MCTS formulation has established methods of using learned domain knowledge to speed up computation time. Specifically, prior work in computational game-playing literature has developed many techniques for applying learned domain information to effectively reduce tree breadth and depth ~\cite{coulom2007computing, silver2016mastering}. We apply similar techniques here to achieve similar reduction in computational cost.

\subsection{Applying learned domain knowledge via the policy distribution}

Monte-Carlo methods for tree search function in the following manner. Starting in an initial state, $s_0$ , obtain a collection of sample rollouts, in which one starts with an action, $a$, and continues acting according to a policy. The expected utility of starting in $s_0$, performing action $a$, and continuing thereafter can then be expressed via the expectation:

\begin{equation}
Q\left(s_0,a\right)=\frac{1}{N\left(s_0,a \right)} \cdot \sum_{i=1}v_i
\end{equation}

in which $N\left( s_0, a \right)$ is the number of times action  has been selected from state $s_0$, and the collection $v_1, v_2, \cdots$ is the empirically observed (or simulated) value of arriving in the state following $s_0$, and continuing to act (i.e. rollouts starting from $s_0$ in which $a$ was taken).
	The question, then, is how the sample of rollouts is collected. If one had perfect knowledge about the expected utility, a simple arg-max operation could be performed over actions at each node to reveal the best node straight away. This would represent essentially a straight march from the initial state to a child node of high utility. At the other extreme is flat, uniform sampling (so-called flat Monte Carlo). In practice, although perfect knowledge is not typically achievable, one can do significantly better at focusing action search than uniform-random sampling.
	In the UCT family of algorithms, The UCB1 heuristic is applied to instead focus sampling on the most promising parts of the action space, resulting in better  value estimates, significantly more quickly, with provably optimal regret bounds. For example, prior to the advent of AlphaGo~\cite{silver2016mastering}, Go programs using variants of UCT achieved amateur level skill ~\cite{lee2010special} - a feat far beyond uniform sampling.
	More generally, the action space is focused by sampling from a policy distribution, $p\left(a|s \right)$. As in all problems involving reinforcement learning, this policy must be selected in such a way that balances exploration and exploitation. One typically favors exploration early on, and gradually focuses on the actions that have proved fruitful. UCT adds new nodes to the tree by first iteratively moving down the existing tree, beginning at the root, by selecting the action $a_j$ to maximize the expression:
	
\begin{equation}
Q\left(s,a_j\right) + 2\cdot C \sqrt{\frac{2\cdot ln(n)}{n_j}}
\end{equation}

$Q\left( s, a_j \right)$  is the current estimate of the expected utility of performing $a_j$ in $s$ and then continuing with the optimal policy. $n$ is the number of times the parent node (representing state $s$) has been visited, and $n_j$ is the number of times this child (representing the state reached when $a_j$ is taken) has been visited. There is a large amount of pressure earlier on to choose unexplored actions (when $n_j$ is small), and this pressure decays as actions get visited. 

In~\cite{lee2010special}, an alternative to the above is used that replaces the exploration term with one that employs a learned probability of action:

\begin{equation}
Q\left(s,a_j\right) + C \cdot \frac{P\left(s,a_j\right)}{1 + N\left(s,a_j \right)}
\end{equation}

in which $P\left(s,a_j \right)$ is interpreted as the prior probability that state-action pair $\left(s,a_j \right)$ is selected. As captured in this expression, the selection criteria results in the mode of $P\left(s,a_j \right)$ being selected early on, when the visit count for $a_j$ is low. The weight behind this mode then gradually decays over time, if another action that proves fruitful is discovered. This is precisely the behavior that is desired to be captured and is adopted in this work.
	Here, $P\left(s,a \right)$ is represented in the following way. From the sampled estimates of $Q\left(s,a \right)$ acquired from MCTS trees built using UCT, it is empirically observed that, given a state, $s$, the $Q$ value utility estimates of the resulting states are very unevenly distributed - usually it is a single action in particular that results in a state of outsize estimated utility value.  For this reason, it is reasonable to use a classification model, $f_\theta\left(s\right)$, which represents the probability that $s$ is an outsize-utility state, with $\theta$ representing the parameters of the model. Once $f_\theta\left(s\right)$ is obtained, $P\left( s, a_j \right)$ is set equal to $f_\theta\left( s' \right)$ where $s'$ is the resultant state after performing $a_j$ in state $s$.

\subsection{Data collection and training}

In the computational games literature, data used to learn policy distributions often comes from recorded games, as well as agents playing simulated games against human or software opponents. In contrast, in this work sampled estimates of the value function are obtained from previously built MCTS trees, in which rollouts are sampled via UCT. These sampled estimates are then used to learn $f_\theta\left(s\right)$, the probability that $s$ is a state of outsized-utility. This is done in the following manner:

\textbf{Data generation}
First, a reference collection of unoptimized Apache Spark programs was selected. In this preliminary investigation, three programs were chosen: a program for performing classification via logistic regression ($LogReg$), a program for performing k-means classification via expectation maximization ($KMeans$), and a program for performing basic extract-transform-load operations ($ETL$). For each program, the following process was then executed. At each step of the UCT algorithm, a new node was selected according to the domain-independent UCT heuristic. The new node, representing a slightly altered runnable variant of the original program, was sent to a Spark cluster to execute. Once executed, performance metrics were extracted (from the Spark history server), and processed in order to extract the desired performance metric. In this case, execution time was chosen as the metric of optimization, although one could just as well have selected, memory usage, for example. This value is then used in the back propagation step. For convergence guarantees, UCT requires the individual reward signals to be between 0 and 1 (inclusive), and so a suitable normalization step occurs before back propagation. This process continues until an entire tree is built. An experimental pipeline was created to automate the above workflow.

\textbf{Extracting features from data}
Once each tree is built, each node in the tree contains a sample of real, measured values for expected utility. Thus, for each tree node, a $featurized-state$, $q-value-estimate$ pair is extracted. Features are extracted from program contents in the form of dependency networks between variables ~\cite{allamanis2017learning}. More precisely, we use the hypothesis that program performance in the Spark programming environment is largely impacted by RDDs, how they are related to each other, and what happens to them. With this in mind, features are extracted from program contents in the following manner. First, variable entities that are likely to be RDDs are identified (these are identified as those tokens on which an RDD API call is performed at some point in the program contents). Once the RDDs are assembled in an index, four adjacency matrices are formed, capturing relations between the RDDs:

\begin{itemize}
\item $assign$. An edge is present between entity A and B if A is reachable from B via a series of assignments.
\item $partition$. An edge is present between entity A and B if A is the assignment target of the partitionBy method being called on B.
\item $join$. An edge is present between A and B if a form of A.join(B) or B.join(A) is present.
\item $broadcast$. An edge is present between A and B if A is broadcast to worker machines to execute a broadcast-join with B.
\item Additionally, there are unary relations that are kept track of, such as if an RDD has been cached to memory (such as $persist$).
\end{itemize}

\begin{figure}[ht]
\vskip 0.2in
\begin{center}
\centerline{\includegraphics[scale=0.4]{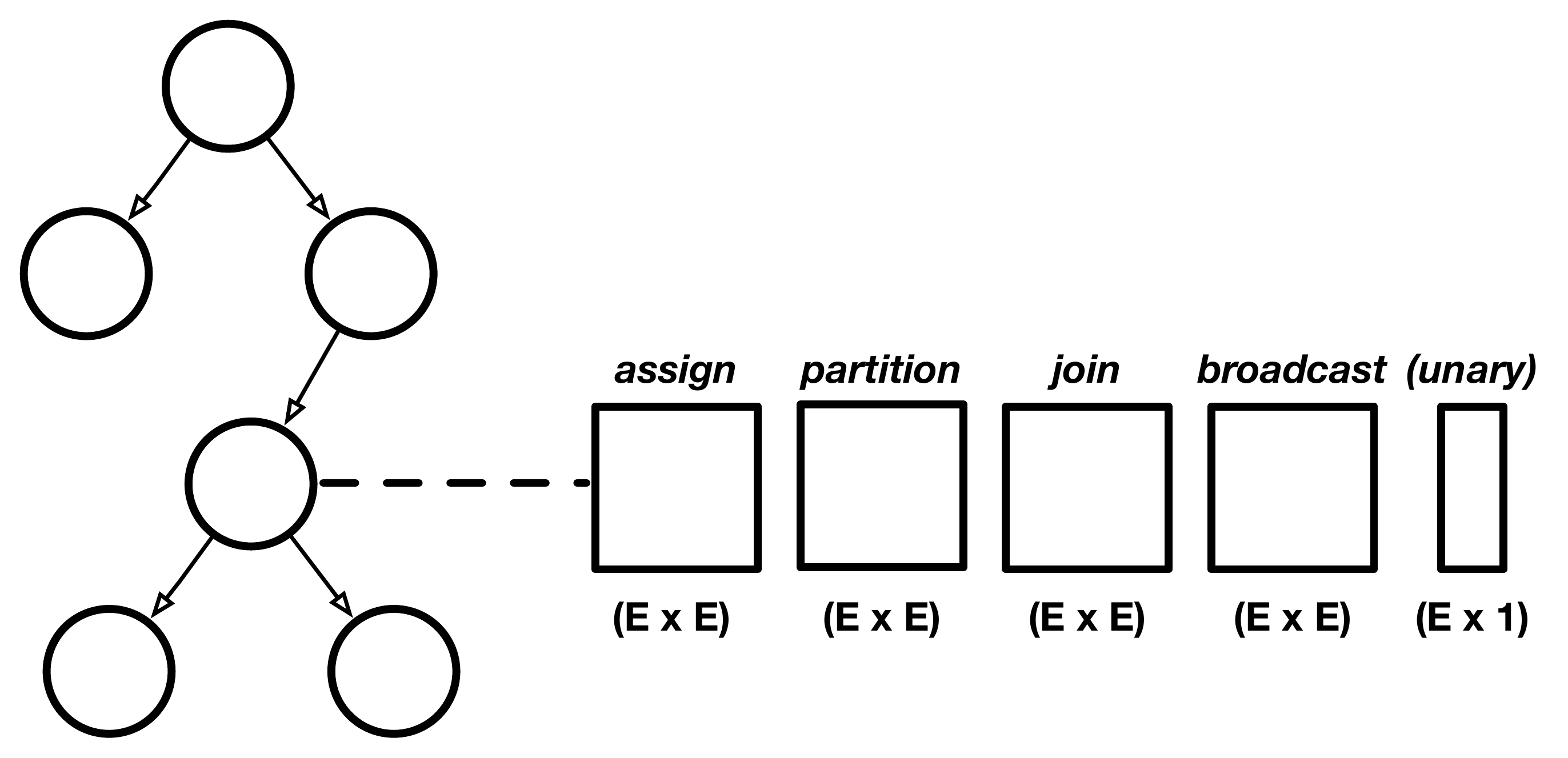}}
\caption{Illustration of how a program's contents, taken from a node in the MCTS tree, is converted into a feature representation. $E$ refers to the number of distinct entities (RDDs) found in the program. The last array is one dimensional, as it contains a single entry for each entity, unlike the binary-relational matrices.}
\label{fig1}
\end{center}
\vskip -0.2in
\end{figure}

Each program is in this manner encoded into a sequence of relational matrices. This is illustrated in figure~\ref{fig1}.

On the left hand side of figure~\ref{fig1}, a MCTS tree is depicted, containing the sequences of alterations made to the original, root program. Each node, representing a (potentially runnable) program, is encoded as raw features, with four matrices representing the program entities' binary relationships, in addition to a one dimensional array, which encodes unary relations.

\begin{figure}[ht]
\vskip 0.2in
\begin{center}
\centerline{\includegraphics[scale=0.4]{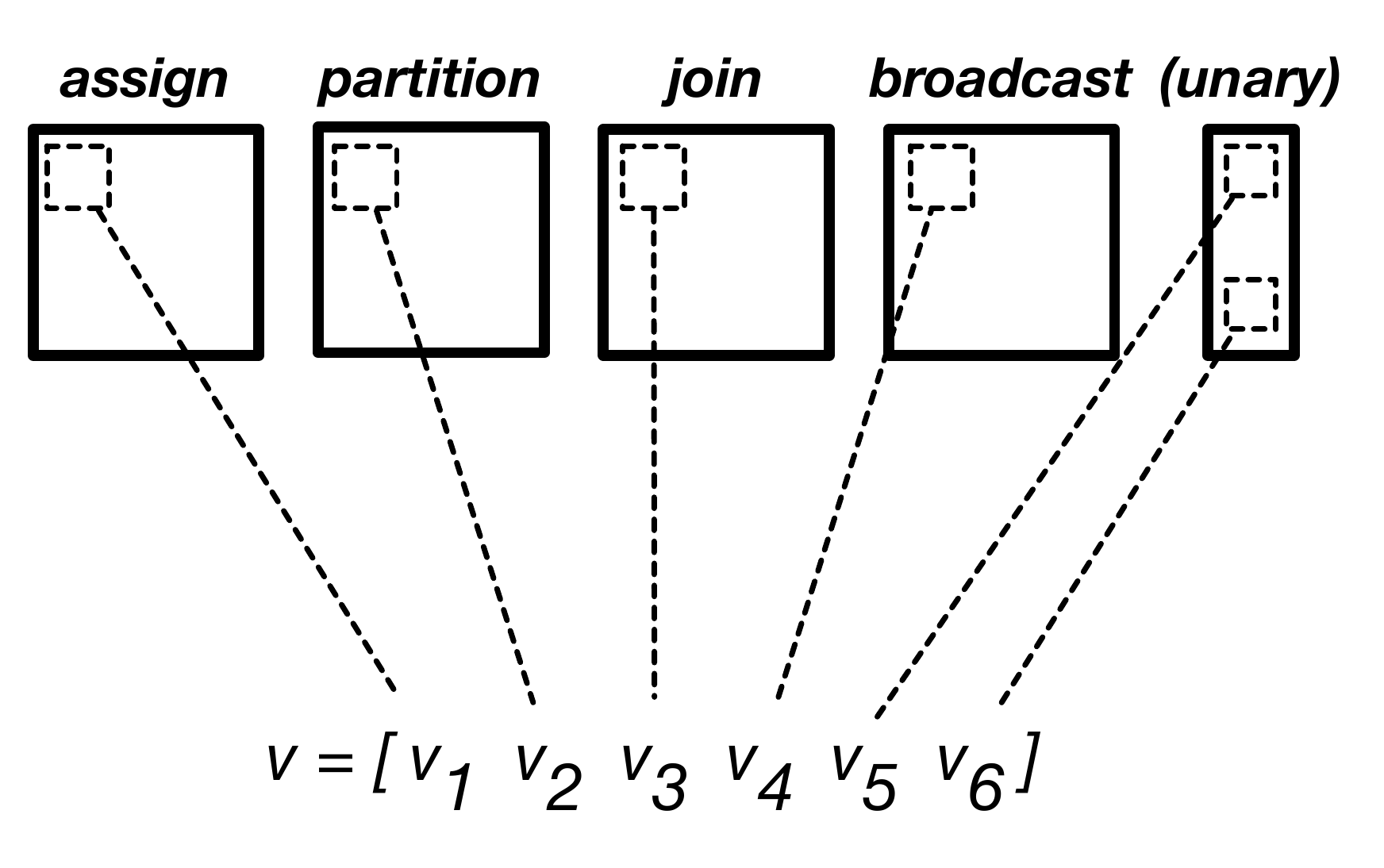}}
\caption{Diagram illustrating how initial features for a program's text contents, represented by a sequence of four binary relational matrices and a one dimensional array, are synthesized into a sequence of vectors of length 6. There is one such vector for each entity-pair in the document, with entities sorted by order of appearance in the document.}
\label{fig2}
\end{center}
\vskip -0.2in
\end{figure}

There is a translational invariance property that is not captured by this, however (e.g.. it is not necessarily significant that precisely the third program entity token is cached). In order to capture this property via a convolutional model, these features are further processed in the following manner. Each document is represented as a sequence of vectors of length six, with one vector for each entity pair. Following the order in which entities appear in the document, a vector is encoded as shown in figure~\ref{fig2}. Specifically, for each type of relation the entity can appear in, a 0 or 1 appears to indicate if it is related to a second entity. The unary array is responsible for two entires in this synthesized feature vector, as there will be one entry for each entity in the pair. In this way, each program state contained in a tree node is represented by a composition of such vectors. 

It should be noted that a tree created via UCT produces data containing an implicit bias. Specifically, it contains sampled estimates for expected utility biased towards more desirable (higher reward) parts of the feature space.

\textbf{The model}
$f_\theta\left(s\right)$ is represented as a neural network with an initial convolutional layer  with 9 filters, with a window size of 10, followed by a max-pool layer. There is a single sigmoid output neuron representing the probability estimate.

\subsection{Using learned policy distribution in search}

Once learned, $f_\theta \left( s\right)$ is incorporated in the action selection heuristic in the following manner. Given the state transition model is deterministic, the state resulting from performing $a_j$ in $s$ is denoted by $T\left(s,a_j \right)$. $P\left( s, a_j \right)$ is thus set equal to $f_\theta \left( T\left(s,a_j \right)\right)$.
	The search process executes in the following manner. Beginning with the root node, $s_0$, enumerate the valid actions, forming the set $\{ a_1, a_2, \cdots, a_k \}$. The action $a_j$ is then chosen that maximizes the expression:
	
\begin{equation}
Q\left(s,a_j\right) + C \cdot \frac{ f_\theta\left(T\left( s,a_j \right) \right) }{1 + N\left(s,a_j \right)}
\end{equation}

This procedure continues, until it produces a leaf node. At that point, the new node, representing a program variation, is sent to the computing cluster for evaluation. Its performance metrics are extracted from downloaded performance logs, and the resulting value is back propagated.
	Back propagation occurs in the following manner. Beginning with the new leaf node, each parent node along the route to the root has its q value estimate and visit count updated. The leaf node has its q value estimate set to the extracted metric (after normalization). This value is sent upwards - the parent node adds the received value to its current q value, increments its visit count by one, and recurses upwards. Optionally, the reward value sent to the parent node can have a decay factor $\gamma \left(0 < \gamma \leq 1 \right)$ applied to it first, the result being to assign a higher reward to taking a favorable action sooner.

\section{Experiments and results}
\label{experiments}

\textbf{Preparation of training data.}
Programs were converted to feature representations via a composition of vectors, with each vector representing the ways in which a pair of program entities are related, as described above. In the training data, each program is paired with a a sampled utility value estimate. Instead of directly learning a utility estimator via a regression model, a binary classifier was learned that estimates the probability that a state is of outsized utility. To facilitate training this model, the sampled utility values in the training data were set to 1 or 0 via an empirically determined cutoff. Training data was then acquired in this manner for the $ETL$, $LogReg$, and $KMeans$ programs. A binary classifier, represented by $f_\theta\left(s\right)$, was then trained. The binary classifier is implemented as a neural network with an initial convolutional layer with 9 filters, with a window size of 10, followed by a max-pool layer. There is a single sigmoid output neuron representing the probability estimate. The network was then trained using cross-entropy loss.

\textbf{Evaluation methodology.}
The hypothesis is that the time required for MCTS to find optimizations for a given program is reduced by using a sampling policy learned from a different program. To evaluate this, the $LogReg$ program is used as a benchmark.
 
	The best performing program alteration discovered in a tree of depth 4 is one in which precisely four alterations have been made from the initially supplied program. Specifically, the data matrix has been partitioned and cached, and the corresponding label vector used in logistic regression has been equivalently partitioned and cached, resulting in a total program speed up from 18 minutes to 4 minutes (on a test logistic regression data set). The reason for the speedup is joins occurring between the data matrix and the label vector for associated records have now been co-located on the same machine, and so the resulting join operation no longer incurs network latency - a major focus of optimization in Apache Spark (and distributed computation more generally).
	
	An optimally performing heuristic would make the optimal choice at each node, marching to the above high-performing target node in a straight line from the root node. Thus, we measure the transferred heuristics according to how close they come to this standard. More precisely, the training data produced from running MCTS with UCT on program A (where program A is one of ETL, KMeans, or LogReg) is used to train a binary classifier  thus producing a sampling policy via the method detailed earlier. This sampling policy is then used on the $LogReg$ program, and the number of steps before the high-performing target node described above is discovered, is recorded. The results are displayed in Table \ref{table:perftable}.
	
\begin{table}[t]
\caption{Performance of MCTS on LogReg after using sampling policy trained from sources.}
\label{table:perftable}
\vskip 0.15in
\begin{center}
\begin{small}
\begin{sc}
\begin{tabular}{lcccr}
\toprule
Policy data source & num steps to $LogReg$ target \\
\midrule
Uniform random & $> 300$\\
$LogReg$ (unaided)   & 120 \\
$LogReg$  & 4 \\
$KMeans$    & 6 \\
$ETL$     & 23\\
$\mathbf{OPTIMUM}$ & $\mathbf{4}$\\
\bottomrule
\end{tabular}
\end{sc}
\end{small}
\end{center}
\vskip -0.1in
\end{table}

In the optimal case, the number of steps is precisely four. As a baseline, the effect of following a uniform random policy (in which the tree policy is to select a uniformly random non-terminal child node, and the sampling policy is to choose an action uniformly at random). These two strategies represent the opposite extremes of perfect domain knowledge, and no domain knowledge. In addition, the number of steps $LogReg$ takes to find the target using domain independent UCT is also included. Ideally, if any useful information has been successfully transferred from another program to $LogReg$, its performance will be somewhere between $LogReg$ unaided and the optimum. Indeed this is the case for using both the $KMeans$ and the $ETL$ data set. Neither performs at the theoretical optimum, but both clearly result in transferring information and have reduced the number of search steps by approximately a factor of 10, compared to $LogReg$ unaided.


\section{Conclusion}
\label{future}

In this paper, an attempt was made to show how the difficult, unscalable process of making changes to the contents of a software program in order to improve its performance can be effectively modeled via decision theoretic frameworks and  monte-carlo tree search in particular. First, this has the initial benefit of being able to handle the delayed-reward, evaluative feedback nature of performance programming. Second, the wealth of literature on using MCTS in computational games shows how MCTS performance can be dramatically improved via learning a sampling policy. 

A method for learning a sampling policy from samples of the utility function produced by following the domain-independent UCT heuristic was proposed. It was then shown how data samples produced from optimizing two separate programs ($ETL$ and $KMeans$), produced a sampling policy that remained effective in a completely different program ($LogReg$). This suggests that the combination of MCTS and neural networks has promising potential in distilling knowledge useful in optimizing a broad range of programs. 

This paper uses Apache Spark as a test bed, but the ideas presented here apply to performance programming more generally.

\printbibliography

\end{document}